\documentclass[journal]{IEEEtran}
\usepackage{cite}
\ifCLASSINFOpdf
\else
\fi

\usepackage{url}
\usepackage{fixltx2e}
\usepackage[]{graphicx}
\usepackage{placeins}
\usepackage{hyperref}
\usepackage{color}
\usepackage{array}
\usepackage{booktabs}
\usepackage{subcaption}
\usepackage{footmisc}
\usepackage[]{moresize}
\usepackage{parskip}
\usepackage[table]{xcolor}
\usepackage[switch]{lineno}
\begin{document}
%\linenumbers
%\title{Convolutional neural networks \\ for automatic segmentation \\ of preterm neonatal and adult MR brain images}
\title{Automatic segmentation of MR brain images \\ with a convolutional neural network}

\author{Pim~Moeskops\textsuperscript{a,b}, Max~A.~Viergever\textsuperscript{a}, Adri\"{e}nne~M.~Mendrik\textsuperscript{a}, Linda~S.~de~Vries\textsuperscript{b}, Manon~J.N.L.~Benders\textsuperscript{b} and~Ivana~I\v sgum\textsuperscript{a}% <-this % stops a space
\\\medskip
\textsuperscript{a}Image Sciences Institute, University Medical Center Utrecht, The Netherlands\\
\textsuperscript{b}Department of Neonatology, University Medical Center Utrecht, The Netherlands
%\thanks{Pim Moeskops (e-mail: p.moeskops@umcutrecht.nl), Max A. Viergever, Adri\"{e}nne M. Mendrik and Ivana I\v sgum are with the Image Sciences Institute, University Medical Center Utrecht, The Netherlands}% <-this % stops a space
%\thanks{Pim Moeskops, Linda S. de Vries and  Manon J.N.L. Benders are with the Department of Neonatology, University Medical Center Utrecht, The Netherlands}% <-this % stops a space
\thanks{This paper has been published in May 2016 as: Moeskops, P., Viergever, M. A., Mendrik, A. M., de Vries, L. S., Benders, M. J., and I\v{s}gum, I. (2016). Automatic segmentation of MR brain images with a convolutional neural network. \textit{IEEE Transactions on Medical Imaging,} 35(5), 1252--1261.}% <-this % stops a space}
}%
\maketitle

\begin{abstract}
Automatic segmentation in MR brain images is important for quantitative analysis in large-scale studies with images acquired at all ages.

This paper presents a method for the automatic segmentation of MR brain images into a number of tissue classes using a convolutional neural network. To ensure that the method obtains accurate segmentation details as well as spatial consistency, the network uses multiple patch sizes and multiple convolution kernel sizes to acquire multi-scale information about each voxel. The method is not dependent on explicit features, but learns to recognise the information that is important for the classification based on training data. The method requires a single anatomical MR image only.

The segmentation method is applied to five different data sets: coronal T\textsubscript{2}-weighted images of preterm infants acquired at 30 weeks postmenstrual age (PMA) and 40 weeks PMA, axial T\textsubscript{2}-weighted images of preterm infants acquired at 40 weeks PMA, axial T\textsubscript{1}-weighted images of ageing adults acquired at an average age of 70 years, and T\textsubscript{1}-weighted images of young adults acquired at an average age of 23 years. The method obtained the following average Dice coefficients over all segmented tissue classes for each data set, respectively: 0.87, 0.82, 0.84, 0.86 and 0.91. 

The results demonstrate that the method obtains accurate segmentations in all five sets, and hence demonstrates its robustness to differences in age and acquisition protocol. 
\end{abstract}


% \fi
\IEEEpeerreviewmaketitle

\section{Introduction}
\label{sec:introduction}
%\IEEEPARstart{T}{his} demo file is intended to serve as a ``starter file''
%for IEEE journal papers produced under \LaTeX\ using
%IEEEtran.cls version 1.8a and later.
%I wish you the best of success.
%\hfill September 17, 2014

\IEEEPARstart{A}{ccurate} automatic brain image segmentation in magnetic resonance (MR) images is a prerequisite for the quantitative assessment of the brain in large-scale studies with images acquired at all ages \cite{Sala04,Dubo08b,Rodr08,Tham10,Hogs13,Li14,Lyal14,Wrig14,Moes15a}. Especially the field of neonatal brain image segmentation has developed rapidly in the last ten years. A popular approach for brain image segmentation is the use of (population specific) atlases\cite{Fisc02,Pras05,Xue07,Weis09,Haba10,Shi10b,Card13,Makr14}, but also pattern recognition methods are used \cite{Wang15,Moes15}, sometimes in combination with an atlas-based approach \cite{Vroo07,Srho13,Anbe13}. To obtain anatomically correct segmentations, these methods need explicit spatial and intensity information. For atlas-based methods spatial information is provided in the form of an atlas which is deformed to match the image at hand. For methods based on pattern recognition spatial information is included in the feature set as distances within an atlas space \cite{Anbe13}, distances to a brain mask \cite{Moes15}, probabilistic results of a previous segmentation step \cite{Wang15,Moes15}, or by imposing anatomical constraints \cite{Wang15}. Intensity information is, in pattern recognition methods, included as a set of features based on (local) intensity, and atlas-based methods are typically performed by matching intensity information between the atlas and target images.

The explicit definition of such spatial and intensity features could be avoided by using convolutional neural networks (CNNs). CNNs have shown to be successful in several computer vision tasks including the recognition of handwriting \cite{LeCu98} and, more recently, the ImageNet challenge, where 1.2 million 2D pictures are classified into 1000 different classes \cite{Kriz12}. In recent years, CNNs also gained popularity in the field of medical image analysis \cite{Pras13,Roth14,Veta15,Roth15,Roth15a,Bar15,Breb15,Zhan15,Ciom15,Wolt15}. In contrast to classical machine learning methods, CNNs do not require a set of hand-crafted features for the classification, but learn sets of convolution kernels that are specifically trained for the classification problem at hand. While classical machine learning methods applied to image segmentation would use e.g. Gaussian or Haar-like kernels to acquire appearance information, CNNs optimise sets of kernels based on the provided training data. In this way, the system can automatically extract information that is relevant for the task. If spatial and intensity information within the image are important to distinguish between classes, this can be learned from the provided information, much like a human observer would recognise objects within a (medical) image.

CNNs have also been used for brain image segmentation. Zhang et al. \cite{Zhan15} presented a method using CNNs for the segmentation of three tissue types: white matter (WM), gray matter (GM), and cerebrospinal fluid (CSF), in MR brain images of 6--8 months old infants, which have low contrast between WM and GM. The method uses 2D patches of a single size from one image plane in T\textsubscript{1}-weighted, T\textsubscript{2}-weighted and fractional anisotropy images as input for a CNN. Prior to the classification, all voxels that do not belong to either of the three tissue types, i.e. skull, cerebellum and brain stem are excluded using in-house tools. De Br\'{e}bisson et al. \cite{Breb15} presented a method for the segmentation of adult T\textsubscript{1}-weighted MR brain images in 134 regions as provided by the MICCAI challenge on multi-atlas labelling \footnote{\url{https://masi.vuse.vanderbilt.edu/workshop2012}}. The method uses multiple parallel networks of 2D patches in orthogonal planes, a 3D patch, and distances to a previous segmentation step. 

Recent MICCAI challenges in neonatal \cite{Isgu15} and adult \cite{Mend15} MR brain image segmentation show that various segmentation methods achieve accurate results, but also that different methods are better at different aspects of brain image segmentation. The best results per tissue type in both the NeoBrainS12 \footnote{\url{http://neobrains12.isi.uu.nl/mainResults.php}} and the MRBrainS13 \footnote{\url{http://mrbrains13.isi.uu.nl/results.php}} challenges were not obtained by a single best performing method. These challenges also show that, despite the overall accurate segmentations achieved by these automatic methods, various inaccuracies are still present.

This paper presents a method for the automatic segmentation of anatomical MR brain images into a number of classes based on a multi-scale CNN. The multi-scale approach allows the method to obtain accurate segmentation details as well as spatial consistency. Unlike previous work that employs CNNs for a brain image segmentation task \cite{Breb15}, the proposed method allows omitting the explicit definition of spatial features. Furthermore, unlike previous work used for brain image segmentation \cite{Zhan15}, the method uses multiple patch and kernel sizes combined. This approach allows the method to learn multi-scale features that estimate both intensity and spatial characteristics. In contrast to using these multiple patch and kernel sizes, other approaches to multi-scale CNNs, used in different applications, provide multi-scale features by directly using the feature maps after the first convolution layer as additional input for a fully connected layer \cite{Serm11,Li12}. 

Additionally, unlike previous work on brain image segmentation, the method is applied to the segmentation of images of developing neonates at different ages as well as young adults and ageing adults, and to coronal as well as axial images. This allows demonstrating that the method is able to adapt to the segmentation task at hand based on training data.

The paper is organised as follows. Section \ref{sec:method} provides a general description of the method. Section \ref{sec:data} describes the five test data sets that were used. Section \ref{sec:preprocessing} describes the preprocessing that was performed. Section \ref{sec:parameters} provides the parameter settings of the CNN that were determined experimentally with a subset of the data. Section \ref{sec:experiments} lists the evaluation experiments and their results. Section \ref{sec:othermethods} compares the obtained results with the results in the recent literature and in the NeoBrainS12 and MRBrainS13 challenges. Section \ref{sec:multiscale} illustrates the influence of the multi-scale approach, and Section \ref{sec:singleplane} illustrates the influence of the single-plane approach. Finally, Section \ref{sec:discussion} discusses the results.

\begin{figure}[!t]
\centering
\includegraphics[trim=5mm 15mm 15mm 5mm,clip,width=0.45\textwidth]{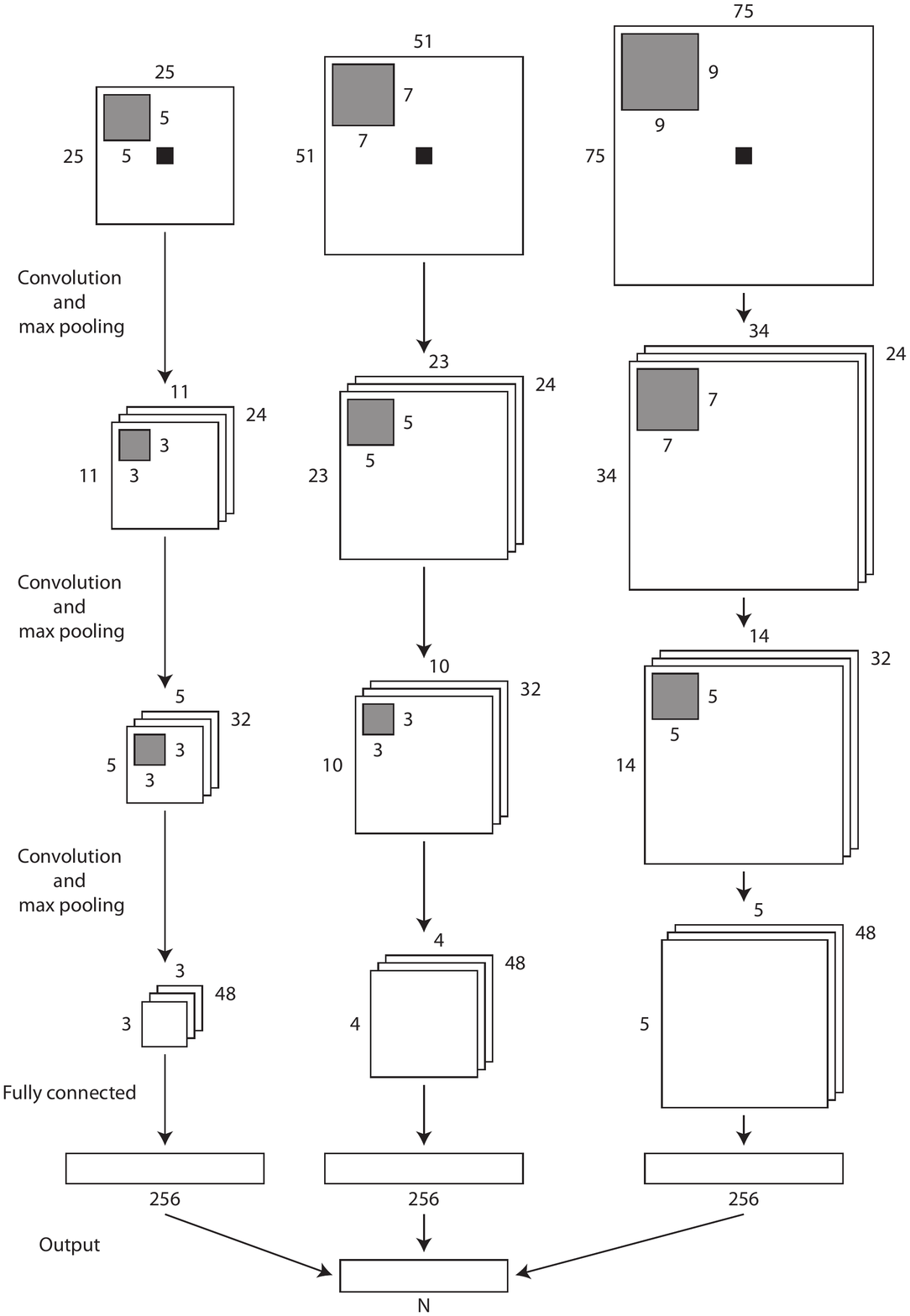} 
\caption{Schematic overview of the convolutional neural network. The number of output classes, N, was set to 9 (8 tissue classes and background) for the neonatal images, to 8 (7 tissue classes and background) for the ageing adult images, and to 7 (6 tissue classes and background) for the young adult images. After the third convolution layer, max-pooling is only performed for the two largest patch sizes.} 
\label{fig:network}
\end{figure}

\section{Method}
\label{sec:method}
Each voxel in the image is classified to one of the brain tissue classes using a CNN. Information about each voxel is provided in the form of image patches where the voxel of interest is in the centre. To allow the method to use multi-scale information about each voxel, multiple patch sizes are used. The larger scales implicitly provide spatial information, because they are large enough to recognise where in the image the voxel is located, while the smaller scales provide detailed information about the local neighbourhood of the voxel. For each of these patch sizes, different kernel sizes are trained, i.e. larger kernel sizes are used for larger patches. This multi-scale approach allows the network to incorporate local details as well as global spatial consistency.

For each of the patch sizes, a separate network branch is used; only the output layer is shared. This allows the weights and biases in the CNN to be specifically optimised for each patch size and corresponding kernel size. Multiple convolution layers are used; after each convolution, the resulting images are sub-sampled by extracting the highest responses with max-pooling. While the dimensions of the patches decrease in each layer, the number of kernels that are trained increases.

After the convolution layers, separate fully connected layers are used for each input patch size. To perform the final classification, these layers are connected to a single softmax output layer with one node for each class (including a background class). 

Because the number of samples per tissue class might vary between the classes, a defined number of samples is extracted per class from each training image. In this way, the number of samples used in the training procedure is the same for each class, which avoids a bias towards larger tissue classes. To provide the network with as many training samples as possible, in every epoch, a new random selection of samples is made and provided to the network in a randomised order. Rectified linear units \cite{Nair10} are used for all nodes because of their speed in training CNNs \cite{Kriz12}. Drop-out \cite{Sriv14} is used on the fully connected layers to decrease the effect of overfitting on the training set. Mini-batch learning and RMSprop \cite{Tiel12} are used to train the network and cross-entropy is used as cost function to optimise the weights and biases. 

A schematic of the CNN that is used in this study is shown in Figure \ref{fig:network}. The parameters that are used in the experiments are described in Section \ref{sec:parameters}.

\begin{table*}[!t]
\scriptsize
\renewcommand{\arraystretch}{1.5}
\caption{Acquisition parameters for the images used in this paper.}
%\rowcolors{2}{gray!25}{white}
\label{tab:acquisition}
\centering
\begin{tabular}{|l|l l l l l|}
\hline
%\rowcolor{gray!50}
& \textbf{Cor. 30 wks} & \textbf{Cor. 40 wks} & \textbf{Ax. 40 wks} & \textbf{Ageing adults} & \textbf{Young adults}\\\hline
Age & 30 weeks PMA & 40 weeks PMA & 40 weeks PMA & 70 years & 23 years \\
Acquisition protocol & Coronal T\textsubscript{2}-weighted &  Coronal T\textsubscript{2}-weighted & Axial T\textsubscript{2}-weighted & Axial T\textsubscript{1}-weighted & Sagittal T\textsubscript{1}-weighted \\
Number of images & 10 & 5 & 7 & 20 & 15\\
Reconstruction matrix & 384 $\times$ 384 $\times$ 50 & 512 $\times$ 512 $\times$ 110 & 512 $\times$ 512 $\times$ 50 & 240 $\times$ 240 $\times$ 48 & 256 $\times$ 256 $\times$ (261--334)\\
Reconstructed voxel sizes [mm\textsuperscript{3}] & 0.34 $\times$ 0.34 $\times$ 2.0 & 0.35 $\times$ 0.35 $\times$ 1.2 & 0.35 $\times$ 0.35 $\times$ 2.0 & 0.96 $\times$ 0.96 $\times$ 3.0 & 1.0 $\times$ 1.0 $\times$ 1.0\\\hline
\end{tabular}
\end{table*}

\begin{figure*}[!t]
\centering
\raisebox{-0.5\height}{\includegraphics[trim=25mm 25mm 25mm 10mm,clip,height=31mm]{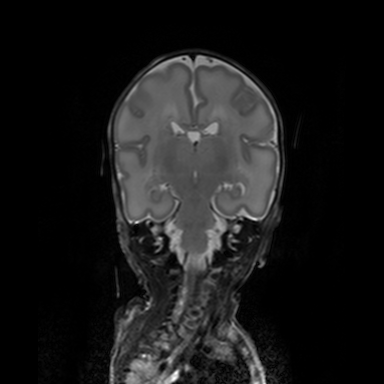}}
\raisebox{-0.5\height}{\includegraphics[trim=10mm 10mm 10mm 10mm,clip,height=35mm]{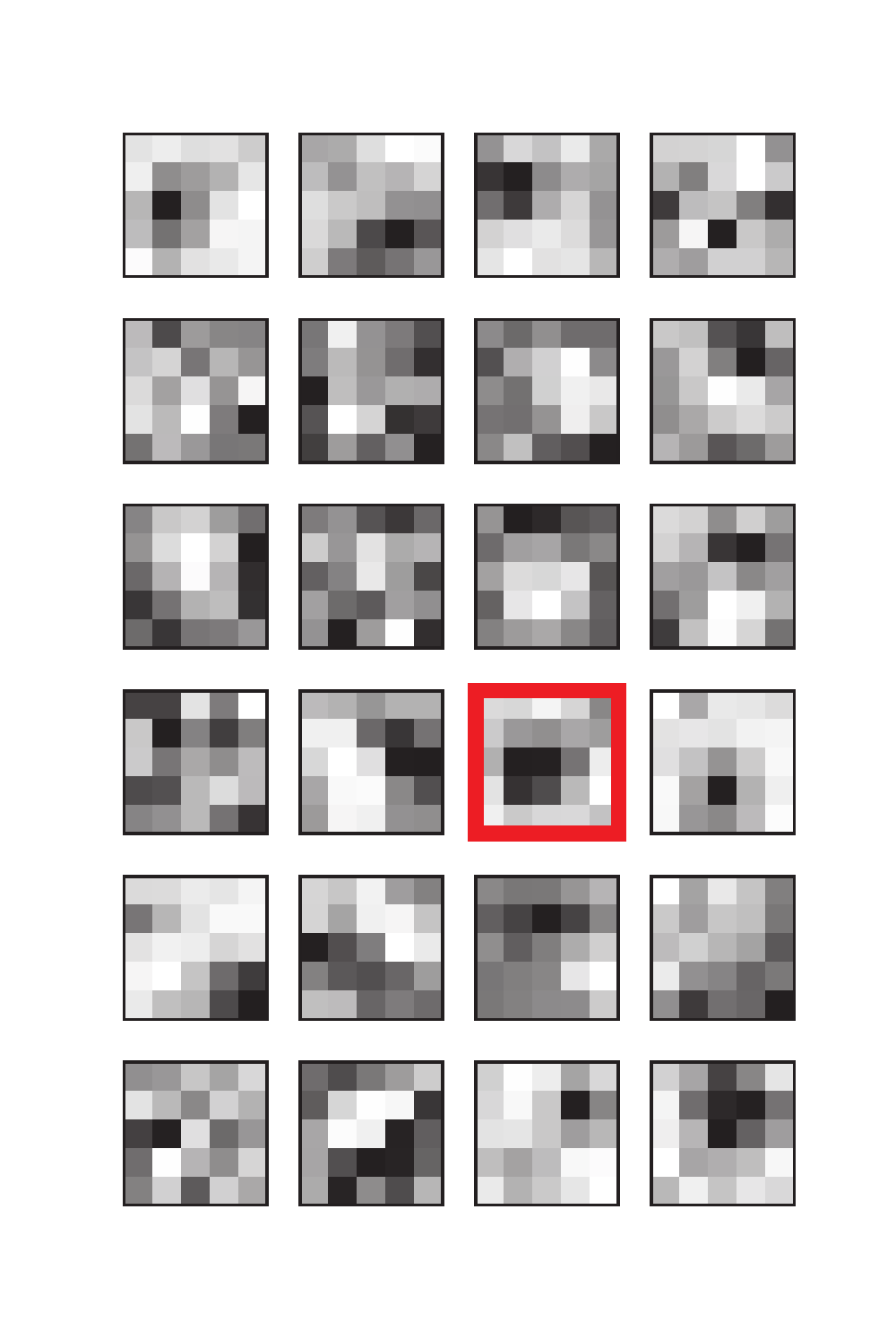}}
\raisebox{-0.5\height}{\includegraphics[trim=25mm 25mm 25mm 10mm,clip,height=31mm]{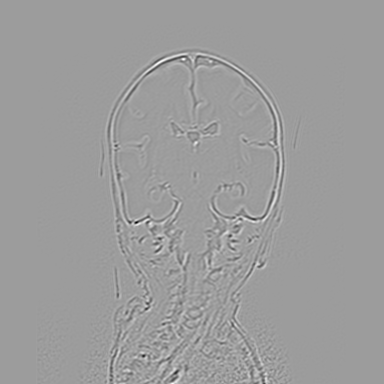}}
\raisebox{-0.5\height}{\includegraphics[trim=10mm 10mm 10mm 10mm,clip,height=35mm]{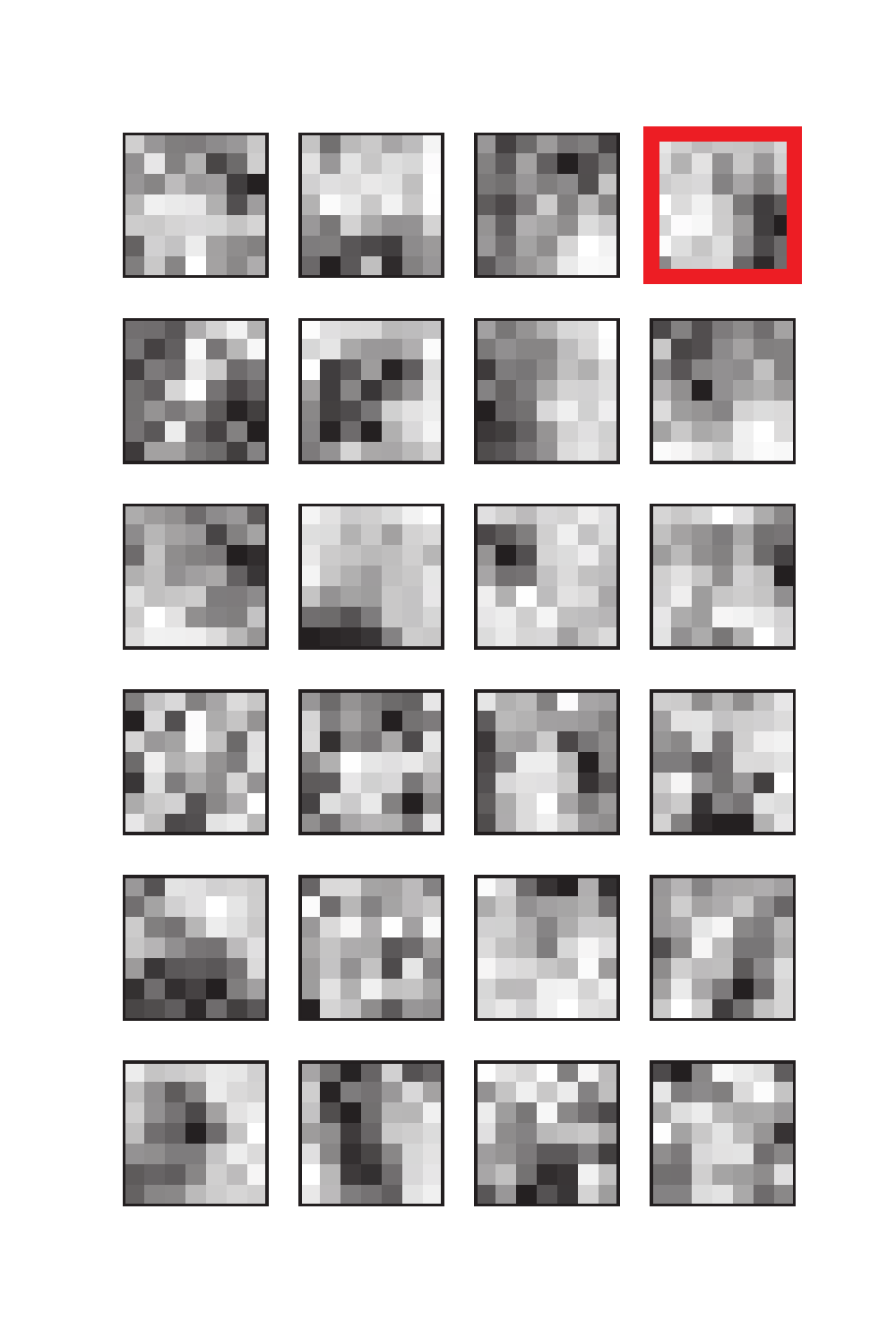}} 
\raisebox{-0.5\height}{\includegraphics[trim=25mm 25mm 25mm 10mm,clip,height=31mm]{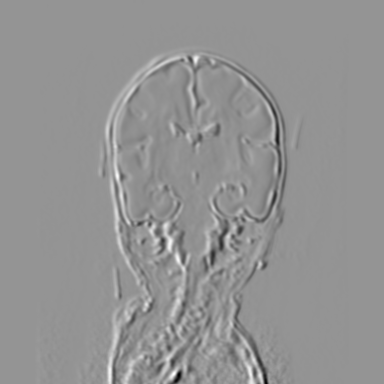}}
\raisebox{-0.5\height}{\includegraphics[trim=10mm 10mm 10mm 10mm,clip,height=35mm]{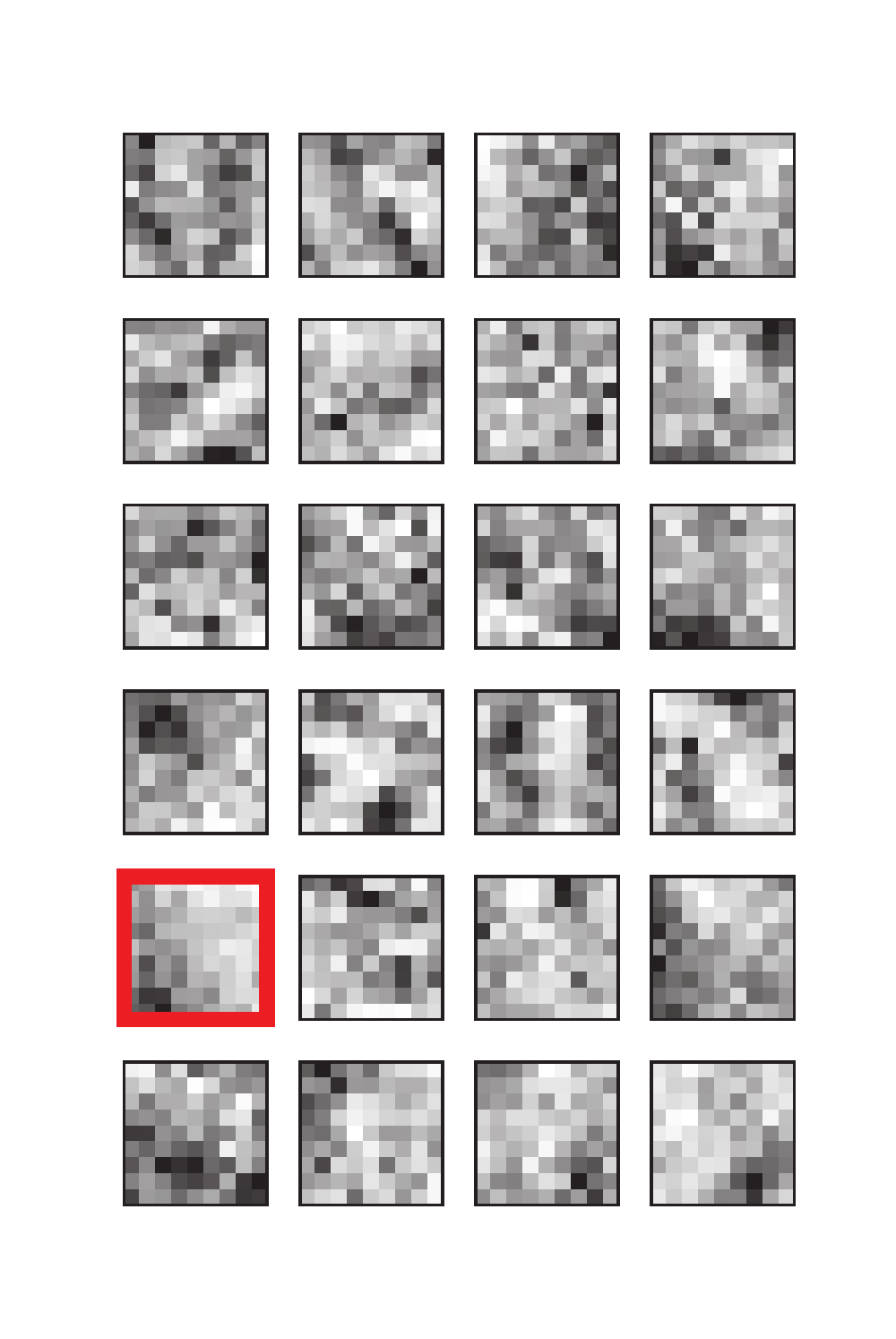}} 
\raisebox{-0.5\height}{\includegraphics[trim=25mm 25mm 25mm 10mm,clip,height=31mm]{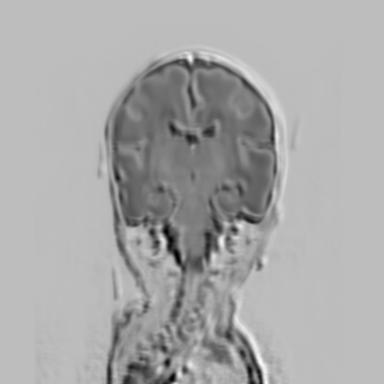}}
\caption{Trained convolution kernels in the first layer after 10 epochs using the 5 training images acquired at 30 weeks PMA, and the kernels indicated in red applied to a test image. From left to right: the T\textsubscript{2}-weighted test image, the kernels of 5 $\times$ 5 voxels, the image convolved with the indicated 5 $\times$ 5 kernel, the kernels of 7 $\times$ 7 voxels, the image convolved with the indicated 7 $\times$ 7 kernel, the kernels of 9 $\times$ 9 voxels, and the image convolved with the indicated 9 $\times$ 9 kernel.}
\label{fig:kernels}
\end{figure*}

\begin{figure*}[!t]
\centering
\includegraphics[trim=55mm 20mm 55mm 25mm,clip,width=0.95\textwidth]{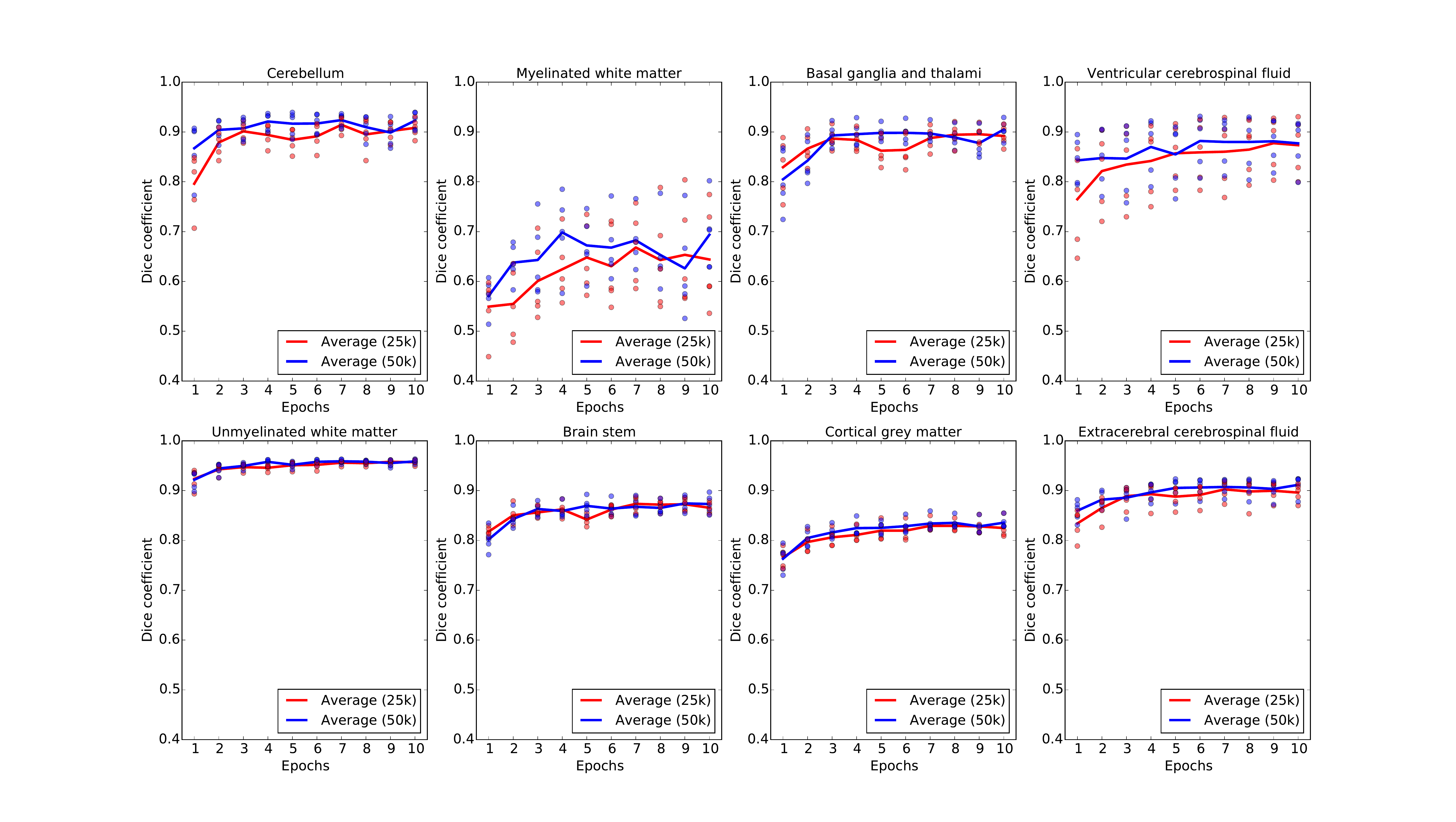} 
\caption{Dice coefficients (automatic vs. reference segmentation volumes) of the 5 test images acquired at 30 weeks PMA for each training epoch when 25~000 (red) and 50~000 (blue) training samples were randomly selected from each class of each of the 5 training images acquired at 30 weeks PMA.}
\label{fig:DiceEpochs}
\end{figure*}

\begin{figure*}[!t]
\centering
\includegraphics[trim=25mm 25mm 25mm 20mm,clip,height=38mm]{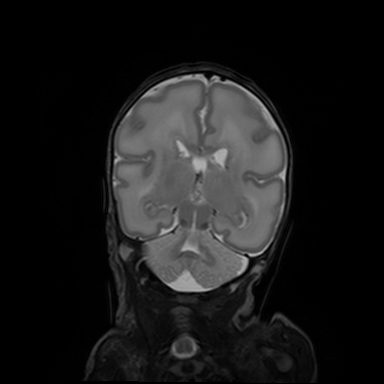}  
\includegraphics[trim=30mm 50mm 30mm 15mm,clip,height=38mm]{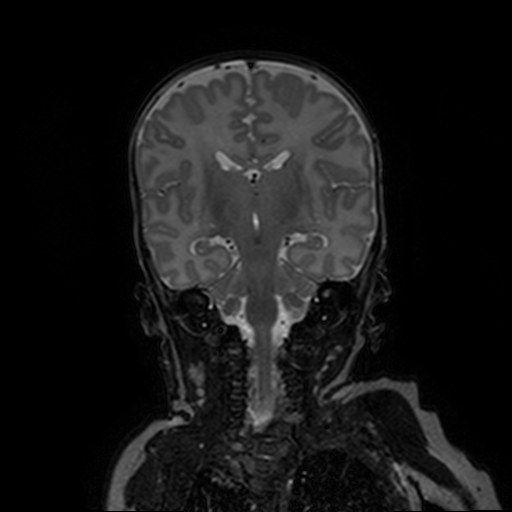}  
\includegraphics[trim=40mm 25mm 40mm 25mm,clip,height=38mm]{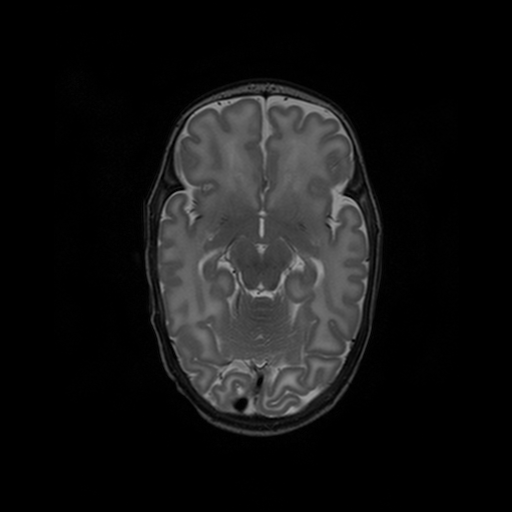} 
\includegraphics[trim=10mm 0mm 10mm 5mm,clip,height=38mm]{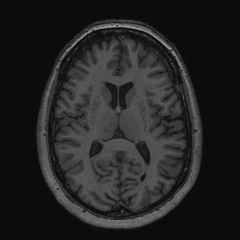}
\includegraphics[trim=10mm 20mm 17mm 10mm,clip,height=38mm]{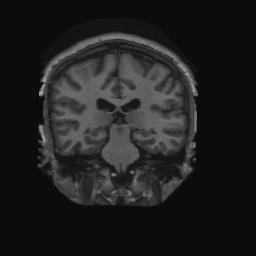}\\\smallskip
\includegraphics[trim=25mm 25mm 25mm 20mm,clip,height=38mm]{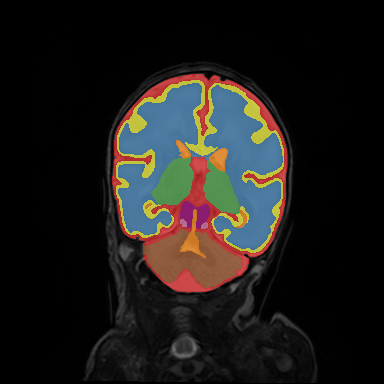}
\includegraphics[trim=30mm 50mm 30mm 15mm,clip,height=38mm]{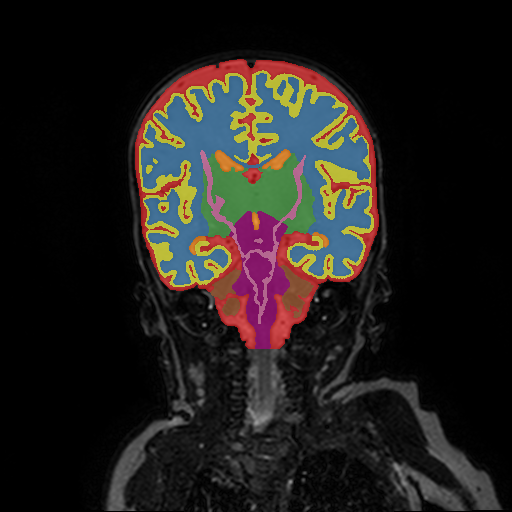}
\includegraphics[trim=40mm 25mm 40mm 25mm,clip,height=38mm]{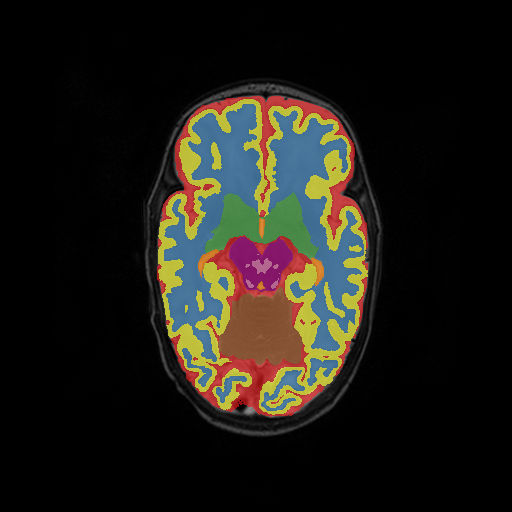}
\includegraphics[trim=10mm 0mm 10mm 5mm,clip,height=38mm]{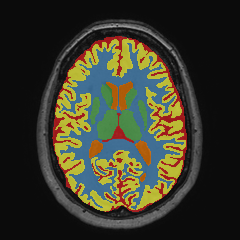}
\includegraphics[trim=10mm 20mm 17mm 10mm,clip,height=38mm]{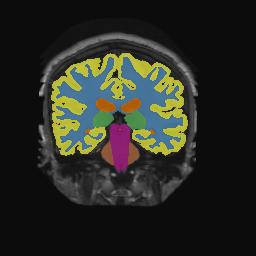}\\\smallskip
\includegraphics[trim=25mm 25mm 25mm 20mm,clip,height=38mm]{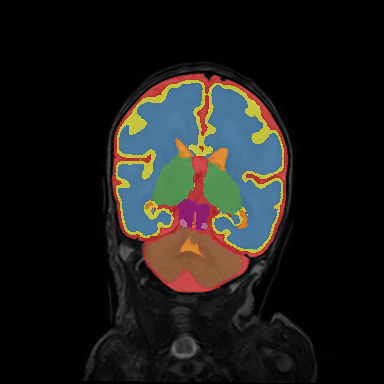}
\includegraphics[trim=30mm 50mm 30mm 15mm,clip,height=38mm]{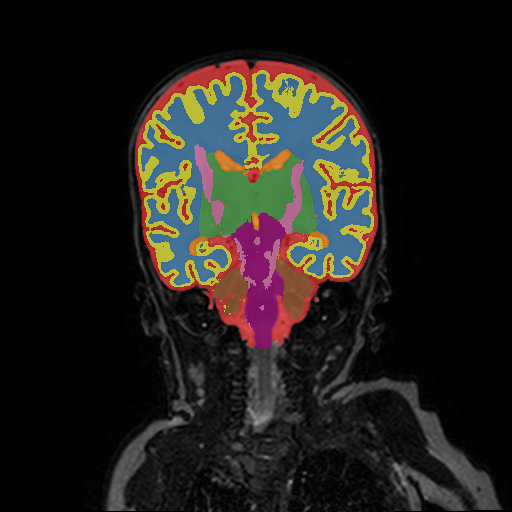}
\includegraphics[trim=40mm 25mm 40mm 25mm,clip,height=38mm]{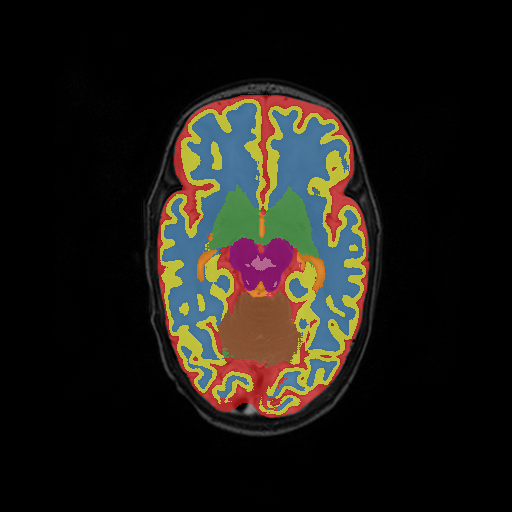}
\includegraphics[trim=10mm 0mm 10mm 5mm,clip,height=38mm]{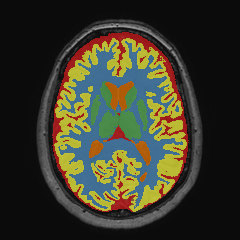}
\includegraphics[trim=10mm 20mm 17mm 10mm,clip,height=38mm]{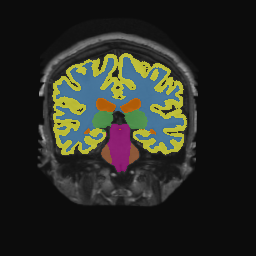}\\
\caption{Segmentation results for CB (brown), mWM (pink), BGT (green), vCSF (orange), (u)WM (blue), BS (purple), cGM (yellow), eCSF (red) in coronal images acquired at 30 weeks PMA (first column), coronal images acquired at 40 weeks PMA (second column), axial images acquired at 40 weeks PMA (third colum), axial images of ageing adults (fourth colum), and images of young adults in the coronal plane (fifth column). The T\textsubscript{2}- or T\textsubscript{1}-weighted image is shown at the top, the manual segmentation in the middle, and the automatic segmentation at the bottom. The images are scaled separately and therefore do not show differences in size.}
\label{fig:segmentations}
\end{figure*}
 
\begin{table*}[!t]
\scriptsize
\renewcommand{\arraystretch}{1.5}
\caption{Results of the presented method in terms of Dice coefficients and mean surface distances (mean $\pm$ standard deviation). The results are listed for (from top to bottom): (1) the coronal images acquired at 30 weeks PMA where 5 training images and 5 test images were used, (2) the 5 coronal images acquired at 30 weeks PMA that are test images in the NeoBrainS12 (NBS12) challenge in leave-one-subject-out (LOSO) cross-validation with all 10 images, (3) the 5 coronal images acquired at 40 weeks PMA in LOSO cross-validation, (4) the 7 axial images acquired at 40 weeks PMA in LOSO cross-validation, (5) the 5 axial images acquired at 40 weeks PMA that are test images in the NBS12 challenge in LOSO cross-validation with all 7 images, (6) the images of ageing adults from the MRBrainS12 challenge where 5 training images and 15 test images were used, (7) the images of young adults from the multi-atlas labelling challenge where 5 training images and 10 test images were used, and (8) the images of young adults from the multi-atlas labelling challenge where 15 training and 20 test images were used.}
\label{tab:diceresults}
\centering
\begin{tabular}{|c|l|l|c c c c c c c c|}
\hline
\multicolumn{11}{|c|}{\textbf{Dice coefficient}}\\\hline
& \textbf{Image set} & \textbf{Experiment} & \textbf{CB} & \textbf{mWM} & \textbf{BGT} & \textbf{vCSF} & \textbf{(u)WM} & \textbf{BS} & \textbf{cGM} & \textbf{eCSF}\\\hline
1 & Cor. 30 wks & 5 training, 5 test & 0.92 $\pm$ 0.02 & 0.69 $\pm$ 0.06 & 0.91 $\pm$ 0.02 & 0.88 $\pm$ 0.05 & 0.96 $\pm$ 0.00 & 0.87 $\pm$ 0.02 & 0.84 $\pm$ 0.01 & 0.91 $\pm$ 0.02\\
2 &  & LOSO 10, NBS12 & 0.91 $\pm$ 0.02 & 0.74 $\pm$ 0.06 & 0.89 $\pm$ 0.02 & 0.91 $\pm$ 0.01 & 0.96 $\pm$ 0.00 & 0.88 $\pm$ 0.02 & 0.83 $\pm$ 0.02 & 0.91 $\pm$ 0.02 \\\hline 
3 & Cor. 40 wks & LOSO 5, NBS12 & 0.93 $\pm$ 0.01 & 0.56 $\pm$ 0.05 & 0.86 $\pm$ 0.02 & 0.85 $\pm$ 0.04 & 0.92 $\pm$ 0.00 & 0.78 $\pm$ 0.05 & 0.82 $\pm$ 0.01 & 0.86 $\pm$ 0.02\\\hline 
4 & Ax. 40 wks & LOSO 7 & 0.93 $\pm$ 0.01 & 0.55 $\pm$ 0.05 & 0.91 $\pm$ 0.02 & 0.81 $\pm$ 0.03 & 0.92 $\pm$ 0.01 & 0.84 $\pm$ 0.02 & 0.88 $\pm$ 0.02 & 0.84 $\pm$ 0.04\\
5 &  & LOSO 7, NBS12 & 0.93 $\pm$ 0.01 & 0.56 $\pm$ 0.06 & 0.91 $\pm$ 0.02 & 0.81 $\pm$ 0.03 & 0.93 $\pm$ 0.01 & 0.85 $\pm$ 0.02 & 0.87 $\pm$ 0.02 & 0.83 $\pm$ 0.04\\\hline
6 & Ageing adults & 5 training, 15 test & 0.90 $\pm$ 0.03 &  -  & 0.81 $\pm$ 0.03 & 0.92 $\pm$ 0.02 & 0.88 $\pm$ 0.02 & 0.90 $\pm$ 0.02 & 0.84 $\pm$ 0.01 & 0.76 $\pm$ 0.04 \\\hline
7 & Young adults & 5 training, 10 test & 0.95 $\pm$ 0.01  & - & 0.85 $\pm$ 0.01 & 0.85 $\pm$ 0.04 & 0.94 $\pm$ 0.01 & 0.92 $\pm$ 0.01 & 0.91 $\pm$ 0.01 & - \\
8 &  & 15 training, 20 test & 0.95 $\pm$ 0.01 & - & 0.86 $\pm$ 0.02 & 0.86 $\pm$ 0.05 & 0.93 $\pm$ 0.02 & 0.93 $\pm$ 0.01 & 0.91 $\pm$ 0.03 & - \\\hline\hline
\multicolumn{11}{|c|}{\textbf{Mean surface distance [mm]}}\\\hline
& \textbf{Image set} & \textbf{Experiment} & \textbf{CB} & \textbf{mWM} & \textbf{BGT} & \textbf{vCSF} & \textbf{(u)WM} & \textbf{BS} & \textbf{cGM} & \textbf{eCSF}\\\hline
1 & Cor. 30 wks & 5 training, 5 test & 0.42 $\pm$ 0.32 & 0.90 $\pm$ 0.70 & 0.49 $\pm$ 0.20 & 0.23 $\pm$ 0.07 & 0.12 $\pm$ 0.01 & 0.29 $\pm$ 0.02 & 0.13 $\pm$ 0.02 & 0.10 $\pm$ 0.02 \\
2 &  & LOSO 10, NBS12 & 0.40 $\pm$ 0.11 & 0.59 $\pm$ 0.34 & 0.49 $\pm$ 0.16 & 0.22 $\pm$ 0.08 & 0.12 $\pm$ 0.01 & 0.25 $\pm$ 0.08 & 0.11 $\pm$ 0.01 & 0.10 $\pm$ 0.02 \\\hline 
3 & Cor. 40 wks & LOSO 5, NBS12 & 0.72 $\pm$ 0.32 & 0.94 $\pm$ 0.44 & 0.96 $\pm$ 0.50 & 0.45 $\pm$ 0.11 & 0.15 $\pm$ 0.01 & 0.87 $\pm$ 0.56 & 0.12 $\pm$ 0.01 & 0.17 $\pm$ 0.02 \\\hline 
4 & Ax. 40 wks & LOSO 7 & 1.04 $\pm$ 0.38 & 0.82 $\pm$ 0.30 & 0.49 $\pm$ 0.20 & 0.67 $\pm$ 0.62 & 0.13 $\pm$ 0.02 & 0.39 $\pm$ 0.12 & 0.11 $\pm$ 0.02 & 0.19 $\pm$ 0.04 \\
5 &  & LOSO 7, NBS12 & 1.14 $\pm$ 0.30 & 0.68 $\pm$ 0.24 & 0.46 $\pm$ 0.19 & 0.43 $\pm$ 0.14 & 0.12 $\pm$ 0.02 & 0.35 $\pm$ 0.08 & 0.11 $\pm$ 0.02 & 0.19 $\pm$ 0.05 \\\hline
6 & Ageing adults & 5 training, 15 test  & 2.01 $\pm$ 2.10 &  -  & 0.67 $\pm$ 0.17 & 0.43 $\pm$ 0.67 & 0.28 $\pm$ 0.05 & 2.41 $\pm$ 2.21 & 0.23 $\pm$ 0.02 & 0.48 $\pm$ 0.10 \\\hline
7 & Young adults & 5 training, 10 test & 1.17 $\pm$ 0.80 & - & 0.75 $\pm$ 0.07 & 0.52 $\pm$ 0.12 & 0.21 $\pm$ 0.05 & 0.64 $\pm$ 0.28 & 0.28 $\pm$ 0.06 & - \\
8 &  & 15 training, 20 test & 0.61 $\pm$ 0.25  & - & 0.72 $\pm$ 0.23 & 0.52 $\pm$ 0.25 & 0.28 $\pm$ 0.17 & 0.46 $\pm$ 0.19 & 0.39 $\pm$ 0.23 & -\\\hline
\end{tabular}
\end{table*}

\begin{table*}[!t]
\tiny
\renewcommand{\arraystretch}{1.3}
\caption{Average Dice coefficients reported in the recent literature on neonatal brain image segmentation and the best results per tissue class in the NeoBrainS12 and MRBrainS13 challenges. It should be noted that the results from the literature are as reported and therefore evaluated on different data sets. The results can therefore not be directly compared but can only provide an indication. Results that are not evaluated or provided in the listed paper are indicated with a '-'. The results of the proposed method are listed in Table \ref{tab:diceresults}.}
\label{tab:diceresultsliterature}
\centering
\begin{tabular}{|l|l|l|c|c c c c c c c c|c c c |}
\hline
\textbf{Source} & \textbf{Acquisition} & \textbf{Age} & \textbf{Test images} & \textbf{CB} & \textbf{mWM} & \textbf{BGT} & \textbf{vCSF} & \textbf{uWM} & \textbf{BS} & \textbf{cGM} & \textbf{eCSF} & \textbf{WM} & \textbf{GM} & \textbf{CSF} \\
\hline
Gui et al., 2012 \cite{Gui12} & 3T coronal T\textsubscript{1} and T\textsubscript{2} & 40 weeks PMA & 10 & 0.87 & 0.78 & 0.88 & - & 0.94 & 0.90 & 0.92 & - & - & - & 0.84 \\
Cardoso et al., 2013 \cite{Card13} & 1.5T T\textsubscript{1} & 40 weeks PMA & 5 & 0.87 & - & 0.84 & 0.94 & 0.91 & 0.74 & 0.76 & - & - & - & - \\
Makropoulos et al., 2014 \cite{Makr14} & 3T axial T\textsubscript{1} and T\textsubscript{2} & 40 weeks PMA & 20 & 0.93 & - & 0.84 & 0.84 & - & 0.92 & - & - & - & - & - \\
Wang et al., 2015 \cite{Wang15} & 3T isotropic T\textsubscript{1}, T\textsubscript{2} and FA & 40 weeks PMA & - &  - &  - & - & - & - &  -  & - & - & 0.92 & 0.89 & 0.84 \\
Moeskops et al., 2015 \cite{Moes15} & 3T coronal T\textsubscript{2} &  30 weeks PMA & 5 & - & - & - &  -  & 0.95 &  -  & 0.81 & 0.89 & - & - & - \\
& 3T coronal T\textsubscript{2} & 40 weeks PMA & 5 &  - &  - & - &  -  &  0.92 &  -  & 0.80 & 0.85 & - & - & -  \\\hline 
NeoBrainS12 & 3T axial T\textsubscript{1} and T\textsubscript{2} & 40 weeks PMA & 5 &  0.92 &  0.47 & 0.92 & 0.86 & 0.90 & 0.83 & 0.86 & 0.79 & 0.92 & - & 0.80 \\
 & 3T coronal T\textsubscript{1} and T\textsubscript{2} & 30 weeks PMA & 5 &  0.88 & 0.69 & 0.84 & 0.88 & 0.91 & 0.76 & 0.71 & 0.83 & 0.90 & - & 0.84 \\
 & 3T coronal T\textsubscript{1} and T\textsubscript{2} & 40 weeks PMA & 5 & 0.92 & 0.48 & 0.88 & 0.84 & 0.89 & 0.75 & 0.77 & 0.77 & 0.84 & - & 0.79 \\\hline
MRBrainS13 & 3T axial T\textsubscript{1}, T\textsubscript{1} IR and T\textsubscript{2} FLAIR & 70 years & 15 & - & - & - & - & - & - & - & - & 0.89 & 0.86 & 0.84 \\\hline
\end{tabular}
\end{table*}

\begin{table*}[!t]
\scriptsize
\renewcommand{\arraystretch}{1.3}
\caption{Comparison of the proposed method on the NeoBrainS12 challenge. The results are ranked according to the average Dice coefficient over all 8 tissue classes. Therefore, only methods that have segmented the full set of 8 tissue classes are included in this comparison. The average was computed over the 3 test images that were segmented by each of the methods. \textsuperscript{*}Note that because no training data was provided for the coronal images acquired at 40 weeks PMA, the average over the 3 test images in the LOSO cross-validation experiment (Table \ref{tab:diceresults}: Experiment 3) is listed.}
\label{tab:comparisonneobrains}
\centering
\begin{tabular}{|l|c|l|c|c|l|c|l|c|c|l|c|l|c|}
\cline{1-4}\cline{6-9}\cline{11-14}
\textbf{Image set} & \textbf{Rank} & \textbf{Method} & \textbf{Dice} & & \textbf{Image set} & \textbf{Rank} & \textbf{Method} & \textbf{Dice} & & \textbf{Image set} & \textbf{Rank} & \textbf{Method} & \textbf{Dice}\\\cline{1-4}\cline{6-9}\cline{11-14}
Cor. 30 wks & \textbf{1} & \textbf{Proposed} & \textbf{0.827} & & Ax. 40 wks & \textbf{1} & \textbf{Proposed} & \textbf{0.805} & & Cor. 40 wks & \textbf{1} & \textbf{Proposed} & \textbf{0.819\textsuperscript{*}}\\  
& 2 & Picsl\_upenn & 0.737 & & & 2 & DTC & 0.789 & & & 2 & DTC & 0.765 \\
& 3 & UCL & 0.728 & & & 3 & UCL & 0.756 & & & 3 & UCL & 0.680\\
& 4 & DTC & 0.721 & & & 4 & Picsl\_upenn & 0.737  & & & 4 & Picsl\_upenn & 0.649 \\\cline{1-4}\cline{6-9}\cline{11-14}
\end{tabular}
\end{table*}

\section{Data}
\label{sec:data}
The method was applied to the segmentation of 3 different sets of volumetric T\textsubscript{2}-weighted MR brain images of preterm infants and 2 sets of volumetric T\textsubscript{1}-weighted MR brain images of adults, hence, 5 different sets of images in total. Because of the inverted contrast between WM and GM in neonatal images as compared with adult images, T\textsubscript{2}-weighted images of neonates  provide a similar contrast between these tissue types as T\textsubscript{1}-weighted images of adults.

Basic information about the images is listed in Table \ref{tab:acquisition}.

\subsection{Neonatal images}
\label{sec:neonatalimages}

In total, 22 images of preterm infants were acquired on a Philips Achieva 3T scanner in accordance with standard clinical practice in the neonatal intensive care unit of the University Medical Center Utrecht (UMCU), The Netherlands. Permission from the medical ethical review committee of the UMCU and informed parental consent were obtained. 10 coronal images were acquired at an age of 30.9 $\pm$ 0.6 (mean $\pm$ standard deviation) weeks postmenstrual age (PMA). For 5 of these patients a coronal follow-up image was acquired at 41.3 $\pm$ 1.3 weeks PMA. Furthermore, 7 axial images (of different patients) were acquired at an age of 41.3 $\pm$ 0.5 weeks PMA. No brain pathology was visible on these images. 

The neonatal images were manually segmented by experts into 8 classes: cerebellum (CB), myelinated white matter (mWM), basal ganglia and thalami (BGT), ventricular cerebrospinal fluid (vCSF), unmyelinated white matter (uWM), brain stem (BS), cortical grey matter (cGM), and extracerebral cerebrospinal fluid (eCSF).

Detailed information about the acquisition and the manual segmentation protocol can be found in the paper by I\v{s}gum et al. \cite{Isgu15} and on the NeoBrainS12 website \footnote{\label{fn:nbs12}\url{http://neobrains12.isi.uu.nl/reference.php}}. I\v{s}gum et al. also evaluated the inter-observer variability for the manual segmentations of the data in the NeoBrainS12 challenge.

\subsection{Ageing adult images}
\label{sec:ageingadultimages}

20 axial images of adults were acquired on a Philips Achieva 3T scanner at an age of 70.5 $\pm$ 4.0 years, as provided by the MRBrainS13 challenge. Permission from the medical ethical review committee of the UMCU was obtained and all participants signed an informed consent form. The patients had a varying degree of atrophy and white matter lesions. 

The ageing adult images were manually segmented by experts into the same classes as the neonatal images. However, because all WM is myelinated in adults, only one WM class was made, resulting in 7 different classes. Possible white matter lesions were included in the WM segmentation.

Detailed information about the acquisition and the manual segmentation protocol can be found in the paper by Mendrik et al. \cite{Mend15} and on the MRBrainS13 website \footnote{\label{fn:mrbs13}\url{http://mrbrains13.isi.uu.nl/details.php}}.

\subsection{Young adult images}
\label{sec:youngadultimages}
15 images from the OASIS project \cite{Marc07} were acquired on a Siemens Vision 1.5T scanner at an age of 23.0 $\pm$ 4.1 years, as provided by the MICCAI challenge on multi-atlas labelling \cite{Land12}. The images were acquired in the sagittal plane with a voxel size of 1.0 $\times$ 1.0 $\times$ 1.25 mm\textsuperscript{3} and were subsequently resized to an isotropic resolution of 1.0 $\times$ 1.0 $\times$ 1.0 mm\textsuperscript{3} \footnote{\url{https://masi.vuse.vanderbilt.edu/workshop2012}}.

The images were manually segmented, in the coronal plane, into 134 classes. To match the tissue definitions used in this paper, these classes were merged into the same tissue classes as used for the other images. Because no eCSF was segmented in these images, this resulted in 6 tissue classes.
 
Detailed information about the acquisition can be found in the paper by Marcus et al. \cite{Marc07} and on the OASIS website \footnote{\url{http://www.oasis-brains.org/}}. Detailed information about the manual segmentation protocol can be found on the NeuroMorphometrics website \footnote{\url{http://neuromorphometrics.org:8080/Seg/}} \footnote{\url{http://braincolor.mindboggle.info/protocols/}}.

\subsection{Evaluation}
\label{sec:evaluation}
The automatic segmentations were evaluated in 3D using the Dice coefficient and the mean surface distance between the manual and automatic segmentations.

\section{Experiments and results}
\label{sec:experimentsandresults}
The parameter settings of the CNN were defined in preliminary leave-one-subject-out experiments with 5 of the images acquired at 30 weeks PMA. This set of images was selected such that it included none of the patients with a follow-up image acquired at 40 weeks PMA. This allowed independent evaluation on the remaining 5 images acquired at 30 weeks PMA as well as on the images of the other 4 test sets by training with representative training data. 

\subsection{Preprocessing}
\label{sec:preprocessing}
Prior to the segmentation, the neonatal images were bias corrected using the method of Likar et al. \cite{Lika01}. The adult images were bias corrected as provided in the MRBrainS13 challenge and the multi-atlas labelling challenge. To limit the number of voxels considered in the classification, brain masks were generated with BET \cite{Smit02a}. In each of the experiments, the samples from the training images were only selected from within the brain mask volumes. For each test image, only voxels within the brain mask volume were considered in the classification. The intensities were scaled such that for all images, the range of intensities within the brain mask was between 0 and 1023.

\subsection{CNN parameters}
\label{sec:parameters}
For all voxels within the brain mask, three in-plane patches with sizes of 25 $\times$ 25, 51 $\times$ 51 and 75 $\times$ 75 voxels are extracted, where the voxel of interest is in the centre. Because of the large slice thickness relative to the in-plane voxel size in 4 of the 5 image sets, orthogonal or 3D patches are not used, i.e. only information from the imaging plane is used for the segmentation of volumetric images. 

A CNN with multiple convolution layers is used; a schematic of the network is shown in Figure \ref{fig:network}. In the first layers 24 kernels are trained for each patch size. For the patches of 25 $\times$ 25 voxels, kernels of 5 $\times$ 5 voxels are used, for the patches of 51 $\times$ 51 voxels, kernels of 7 $\times$ 7 voxels are used, and for the patches of 75 $\times$ 75 voxels, kernels of 9 $\times$ 9 voxels are used. Max-pooling with kernels of 2 $\times$ 2 voxels is performed on the convolved images. The output images are input for the second layer, where 32 kernels of 3 $\times$ 3, 5 $\times$ 5 and 7 $\times$ 7 voxels are used. Again, max-pooling with kernels of 2 $\times$ 2 voxels is performed on the convolved images. In the third layer, 48 kernels of 3 $\times$ 3, 3 $\times$ 3 and 5 $\times$ 5 voxels are used. After this layer, max-pooling is only performed for the two largest patches, because for the smallest patch only an image of 3 $\times$ 3 voxels is left after three convolution layers. The contiguous even-sized max-pooling kernels cannot exactly cover the odd-sized patches. Therefore, before all max-pooling operations, the values are mirrored along the borders. The output of the third layer is input to fully connected layers with 256 nodes for each patch size. These nodes are subsequently connected to the softmax output layer with 9 nodes (8 tissue classes and background) for the neonatal images, 8 nodes (7 tissue classes and background) for the ageing adult images, and 7 nodes (6 tissue classes and background) for the young adult images.  An example of the trained kernels in the first layer and an example of the convolution responses for images acquired at 30 weeks PMA is shown in Figure \ref{fig:kernels}. 

The tissue classes consist of a very different number of voxels in these images. For example, in the images acquired at 30 weeks PMA, mWM consists of about 200 voxels, while uWM consists of 400 000 voxels. Therefore, to better balance the number of training samples of each class a fixed number of samples are extracted per class from each training image. If a class contains less than this number of samples, all samples are used. Multiple epochs are used in the training process, where a new set of random samples is extracted in each epoch. The performance on the test images acquired at 30 weeks PMA for each epoch using 25 000 and 50 000 training samples is shown in Figure \ref{fig:DiceEpochs}. Based on this evaluation, in all experiments, 10 epochs and 50 000 training samples per class from each image are used. 

\subsection{Evaluation experiments}
\label{sec:experiments}
The performance for the images acquired at 30 weeks PMA was evaluated on the independent set of 5 images not included in the parameter estimation, using the other 5 images as training data. For the other image sets, the method was retrained using representative training data. Because of the limited number of available training images, the performance on the coronal and axial images acquired at 40 weeks PMA was evaluated in leave-one-subject-out cross-validation for each set. The average Dice coefficients and mean surface distances for each tissue class are shown per set in Table \ref{tab:diceresults}: Experiments 1, 3, and 4. The segmentation results in one slice of one image of each of the test sets are shown in Figure \ref{fig:segmentations}. 

A subset of the neonatal images used in this paper is used as test data in the NeoBrainS12 challenge. This includes 5 of the coronal images acquired at 30 weeks PMA, all 5 images acquired at 40 weeks PMA, and 5 of the axial images acquired at 40 weeks PMA. Therefore, to allow an \textit{indication} of the performance in comparison with the results in the challenge, the average results for the axial images acquired at 40 weeks are presented separately for the 5 test images in the challenge (Table \ref{tab:diceresults}: Experiment 5). For the coronal images acquired at 30 weeks PMA an additional experiment is done in the same fashion as for the other images: leave-one-subject-out cross-validation with all ten images where the average over the test images in the challenge is listed in Table \ref{tab:diceresults}: Experiment 2.

Given that a larger set of manually annotated images of ageing adults was available, their performance was evaluated using 5 images as training data and 15 images to test the performance (Table \ref{tab:diceresults}: Experiment 6, and Figure \ref{fig:segmentations}). This furthermore allowed (indirect) comparison with the results of the MRBrainS13 challenge, which is performed using the same training and test images, except that only 3 combined tissue classes, namely WM, GM, and CSF are evaluated in the challenge.

Similar to the images of ageing adults, the images of young adults were evaluated using 5 images as training data and 10 images to test the performance (Table \ref{tab:diceresults}: Experiment 7, and Figure \ref{fig:segmentations}). Additionally, the performance on the independent test set of 20 images \cite{Land12} is evaluated using all 15 images as training data (Table \ref{tab:diceresults}: Experiment 8).

\subsection{Comparison with previous methods}\label{sec:othermethods}
For comparison purposes, Table \ref{tab:diceresultsliterature} lists segmentation results in terms of average Dice coefficients reported in the recent literature on neonatal MR brain image segmentation. Note that these methods used different scans that were segmented possibly using different tissue definitions. Hence, these results can only be used as an indication. Furthermore, Table \ref{tab:diceresultsliterature} lists the best results per tissue class at the time of writing in the NeoBrainS12 and MRBrainS13 challenges. Note that in these challenges, the best results for different tissue types are not necessarily obtained by a single method.

To directly compare the method with previous methods, it was also evaluated in the NeoBrainS12 challenge using only the training images provided by the challenge. Table \ref{tab:comparisonneobrains} compares the results of the proposed method with other methods that have segmented the same tissue classes in the same images. Comparison has been performed by averaging the Dice coefficients over all tissue classes in all test images. Detailed comparison per tissue class is available on the NeoBrainS12 website \footnote{\url{http://neobrains12.isi.uu.nl/mainResults.php}}.

\subsection{Multi-scale approach}
\label{sec:multiscale}
\begin{figure*}[!t]
\centering
\begin{subfigure}{.15\textwidth}
\centering
\includegraphics[width=\textwidth]{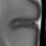}\\\smallskip
\includegraphics[width=\textwidth]{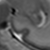}
\caption{Image} 
\end{subfigure}
\begin{subfigure}{.15\textwidth}
\centering
\includegraphics[width=\textwidth]{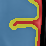}\\\smallskip
\includegraphics[width=\textwidth]{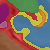}
\caption{Reference} 
\end{subfigure}
\begin{subfigure}{.15\textwidth}
\centering
\includegraphics[width=\textwidth]{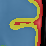}\\\smallskip
\includegraphics[width=\textwidth]{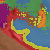}
\caption{25 $\times$ 25} 
\end{subfigure}
\begin{subfigure}{.15\textwidth}
\centering
\includegraphics[width=\textwidth]{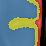}\\\smallskip
\includegraphics[width=\textwidth]{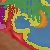}
\caption{51 $\times$ 51} 
\end{subfigure}
\begin{subfigure}{.15\textwidth}
\centering
\includegraphics[width=\textwidth]{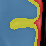}\\\smallskip
\includegraphics[width=\textwidth]{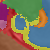}
\caption{75 $\times$ 75} 
\end{subfigure}
\begin{subfigure}{.15\textwidth}
\centering
\includegraphics[width=\textwidth]{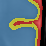}\\\smallskip
\includegraphics[width=\textwidth]{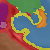}
\caption{3 patch sizes} 
\end{subfigure}
\caption{Segmentation results in a T\textsubscript{2}-weighted image (left) acquired at 30 weeks PMA for the lateral sulcus (top) and the hippocampus (bottom) using (from left to right), manual segmentation, only a patch of 25 $\times$ 25 voxels, only a patch of 51 $\times$ 51 voxels, only a patch of 75 $\times$ 75 voxels, and these 3 patch sizes combined. The tissues are labelled as follows: CB in brown, mWM in pink, BGT in green, vCSF in orange, uWM in blue, BS in purple, cGM in yellow, and eCSF in red.}
\label{fig:zoomedsegmentations}
\end{figure*}

To show the influence of combining different patch sizes in the network, the method was additionally trained using each of the three patch sizes separately. A segmentation result for the lateral sulcus and the hippocampus in an image acquired at 30 weeks PMA using only the patches of 25 $\times$ 25, 51 $\times$ 51 and 75 $\times$ 75 voxels, as well as those three patches combined, is shown in Figure \ref{fig:zoomedsegmentations}. The results in terms of Dice coefficients and mean surface distance over all 5 test images acquired at 30 weeks PMA are shown in Table \ref{tab:diceseparatepatches}: Experiments 1, 3, 5 and 7.

In addition, the proposed multi-scale approach was compared with the multi-scale approach of Sermanet and LeCun \cite{Serm11}. To allow fair comparison with the other experiments, we used the largest patch size, i.e. 75 $\times$ 75 voxels, as input. To acquire multi-scale information in this approach, the output of the first, second and third layers provide combined input for the fully connected layer (Table \ref{tab:diceseparatepatches}: Experiment 6).

\subsection{Single-plane approach}
\label{sec:singleplane}
To motivate the choice of providing the network with patches from the acquisition planes only, the method was additionally evaluated using patches from consecutive slices, patches from the orthogonal planes, and 3D patches. 

First, the method was trained for the images acquired at 30 weeks PMA using three consecutive slices for each of the three patch sizes (25 $\times$ 25, 51 $\times$ 51 and 75 $\times$ 75). Instead of a network with three branches, i.e. one for each of the three patch sizes (Figure \ref{fig:network}), now a network with nine branches, i.e. three for each of the three patch sizes, was used (Table \ref{tab:diceseparatepatches}: Experiment 8). Because of the large anatomical variation between slices due to the large slice thickness in 4 of the 5 image sets, separate branches allow the network to extract the relevant information by optimising dedicated kernels for each of the nine provided input patches. 

Second, the method was trained for the images acquired at 30 weeks PMA using patches from the orthogonal planes (coronal, sagittal and axial) for each of the three patch sizes. Because these images are highly anisotropic, they were first interpolated to isotropic resolution using an interpolation factor of 6 in the out-of-plane direction. This resulted in voxel sizes of 0.34 $\times$ 0.34 $\times$ 0.33. Again, these orthogonal patches of three sizes were input for a network with nine branches (Table \ref{tab:diceseparatepatches}: Experiment 9). 

Third, two networks with interpolated 3D input patches and 3D convolutions were evaluated for the images acquired at 30 weeks PMA, one with input patches of 25 $\times$ 25 $\times$ 25 voxels, and one with input patches of 51 $\times$ 51 $\times$ 51 voxels (Table \ref{tab:diceseparatepatches}: Experiments 2 and 4).

Fourth, because the images of the young adults were already acquired with a (nearly) isotropic resolution, the performance of the network with three orthogonal patches for each of the three patch sizes was evaluated for this set as well. This network was evaluated for the segmentation of the 6 combined tissue classes (Table \ref{tab:diceseparatepatches}: Experiment 11), as well as for the segmentation of all 134 classes. The average Dice coefficient for the segmentation of 134 classes in the 20 test images was 0.7353 overall, 0.7170 for the cortical regions, and 0.7850 for the non-cortical regions. This performance would result in the 9th overall ranking out of 26 participants on the MICCAI challenge on multi-atlas labelling \cite{Land12}.

\begin{table*}[!t]
\scriptsize
\renewcommand{\arraystretch}{1.5}
\caption{Results in terms of Dice coefficients and mean surface distance (mean $\pm$ standard deviation). For the images acquired at 30 weeks PMA the results are shown using patches of (1) 25 $\times$ 25, (2) 25 $\times$ 25 $\times$ 25, (3) 51 $\times$ 51, (4) 51 $\times$ 51 $\times$ 51 and (5) 75 $\times$ 75 voxels separately, (6) 75 $\times$ 75 voxels, where the output of the first, second and third layers are combined input for the fully connected layer, (7) the three 2D patch sizes combined, (8) three consecutive slices for all three patch sizes, and (9) three interpolated orthogonal planes for all three patch sizes. For the images of young adults the results are shown using (10) the three 2D patch sizes combined, and (11) three interpolated orthogonal planes for all three patch sizes. The experiments are performed using 5 images as training data and 5 images to test the performance for the images acquired at 30 weeks PMA, and 5 images as training data and 10 images to test the performance for the images of young adults. Experiment 7 is therefore the same as Experiment 1 of Table \ref{tab:diceresults}, and Experiment 10 is therefore the same as Experiment 7 of Table \ref{tab:diceresults}.}
\label{tab:diceseparatepatches}
\centering
\begin{tabular}{|c|l|l|c c c c c c c c|}
\hline
\multicolumn{11}{|c|}{\textbf{Dice coefficient}}\\\hline
& \textbf{Image set} & \textbf{Experiment} & \textbf{CB} & \textbf{mWM} & \textbf{BGT} & \textbf{vCSF} & \textbf{(u)WM} & \textbf{BS} & \textbf{cGM} & \textbf{eCSF}\\\hline
1 & Cor. 30 wks & 25$\times$25 patch & 0.69 $\pm$ 0.04  & 0.33 $\pm$ 0.13 & 0.70 $\pm$ 0.05 & 0.76 $\pm$ 0.06 & 0.91 $\pm$ 0.02 & 0.58 $\pm$ 0.03 & 0.78 $\pm$ 0.02 & 0.86 $\pm$ 0.02 \\
2 & & 25$\times$25$\times$25 patch & 0.78 $\pm$ 0.03 & 0.60 $\pm$ 0.03 & 0.77 $\pm$ 0.04 & 0.78 $\pm$ 0.02 & 0.92 $\pm$ 0.01 & 0.79 $\pm$ 0.02 & 0.75 $\pm$ 0.02 & 0.85 $\pm$ 0.02 \\
3 & & 51$\times$51 patch & 0.90 $\pm$ 0.02 & 0.57 $\pm$ 0.08  & 0.87 $\pm$ 0.02 & 0.84 $\pm$ 0.04 & 0.94 $\pm$ 0.01 & 0.85 $\pm$ 0.01 & 0.78 $\pm$ 0.02 & 0.88 $\pm$ 0.02\\
4 & & 51$\times$51$\times$51 patch & 0.82 $\pm$ 0.02 & 0.48 $\pm$ 0.14 & 0.77 $\pm$ 0.06 & 0.70 $\pm$ 0.07 & 0.90 $\pm$ 0.01 & 0.77 $\pm$ 0.02 & 0.67 $\pm$ 0.03 & 0.80 $\pm$ 0.01\\
5 & & 75$\times$75 patch & 0.91 $\pm$ 0.01 & 0.62 $\pm$ 0.08 & 0.88 $\pm$ 0.02 & 0.85 $\pm$ 0.06 & 0.94 $\pm$ 0.01 & 0.85 $\pm$ 0.01 & 0.78 $\pm$ 0.02 & 0.88 $\pm$ 0.02 \\
6 & & 75$\times$75, multi-scale & 0.89 $\pm$ 0.02 & 0.61 $\pm$ 0.07 & 0.89 $\pm$ 0.01 & 0.83 $\pm$ 0.07 & 0.94 $\pm$ 0.01 & 0.86 $\pm$ 0.02 & 0.74 $\pm$ 0.03 & 0.86 $\pm$ 0.01 \\
\cline{3-11}
7 & & \textbf{3 patch sizes} & 0.92 $\pm$ 0.02 & 0.69 $\pm$ 0.06 & 0.91 $\pm$ 0.02 & 0.88 $\pm$ 0.05 & 0.96 $\pm$ 0.00 & 0.87 $\pm$ 0.02 & 0.84 $\pm$ 0.01 & 0.91 $\pm$ 0.02 \\\cline{3-11}
8 & & 3 patch sizes, slices & 0.92 $\pm$ 0.01 & 0.68 $\pm$ 0.08 & 0.90 $\pm$ 0.01 & 0.89 $\pm$ 0.05 & 0.96 $\pm$ 0.00 & 0.87 $\pm$ 0.02 & 0.84 $\pm$ 0.01 & 0.91 $\pm$ 0.02 \\
9 & & 3 patch sizes, ortho & 0.93 $\pm$ 0.01 & 0.68 $\pm$ 0.06 & 0.90 $\pm$ 0.01 & 0.89 $\pm$ 0.03 & 0.96 $\pm$ 0.00 & 0.88 $\pm$ 0.02 & 0.83 $\pm$ 0.01 & 0.91 $\pm$ 0.01 \\\hline
10 & Young adults & \textbf{3 patch sizes} & 0.95 $\pm$ 0.01 & - & 0.85 $\pm$ 0.01 & 0.85 $\pm$ 0.04 & 0.94 $\pm$ 0.01 & 0.92 $\pm$ 0.01 & 0.91 $\pm$ 0.01 & - \\\cline{3-11}
11 & & 3 patch sizes, ortho & 0.96 $\pm$ 0.01 & - & 0.88 $\pm$ 0.01 & 0.84 $\pm$ 0.04 & 0.94 $\pm$ 0.01 & 0.93 $\pm$ 0.02 & 0.91 $\pm$ 0.01 & -
\\\hline\hline
\multicolumn{11}{|c|}{\textbf{Mean surface distance [mm]}}\\\hline
& \textbf{Image set} & \textbf{Experiment} & \textbf{CB} & \textbf{mWM} & \textbf{BGT} & \textbf{vCSF} & \textbf{(u)WM} & \textbf{BS} & \textbf{cGM} & \textbf{eCSF}\\\hline
1 & Cor. 30 wks  & 25$\times$25 patch & 8.12 $\pm$ 1.25 & 2.83 $\pm$ 0.94 & 3.82 $\pm$ 0.99 & 1.53 $\pm$ 0.65 & 0.40 $\pm$ 0.07 & 2.09 $\pm$ 0.74 & 0.21 $\pm$ 0.03  & 0.18 $\pm$ 0.04 \\ 
2 & & 25$\times$25$\times$25 patch & 4.76 $\pm$ 0.95 & 1.40 $\pm$ 0.62 & 2.12 $\pm$ 0.55 & 0.85 $\pm$ 0.58 & 0.31 $\pm$ 0.04 & 0.91 $\pm$ 0.09 & 0.21 $\pm$ 0.03 & 0.19 $\pm$ 0.04 \\
3 & & 51$\times$51 patch & 0.47 $\pm$ 0.23 & 1.41 $\pm$ 0.64 & 0.79 $\pm$ 0.18 & 0.33 $\pm$ 0.09 & 0.19 $\pm$ 0.01 & 0.53 $\pm$ 0.21 & 0.19 $\pm$ 0.04 & 0.14 $\pm$ 0.03 \\
4 & & 51$\times$51$\times$51 patch & 1.82 $\pm$ 0.71 & 1.97 $\pm$ 0.70 & 2.66 $\pm$ 1.30 & 1.58 $\pm$ 0.65 & 0.36 $\pm$ 0.02 & 0.88 $\pm$ 0.13 & 0.25 $\pm$ 0.04 & 0.26 $\pm$ 0.05 \\
5 & & 75$\times$75 patch & 0.32 $\pm$ 0.16 & 1.33 $\pm$ 0.53 & 0.63 $\pm$ 0.16 & 0.29 $\pm$ 0.11 & 0.16 $\pm$ 0.01 & 0.36 $\pm$ 0.09 & 0.15 $\pm$ 0.02 & 0.15 $\pm$ 0.02 \\
6 & & 75$\times$75, multi-scale & 0.46 $\pm$ 0.18 & 1.19 $\pm$ 0.51 & 0.56 $\pm$ 0.13 & 0.33 $\pm$ 0.11 & 0.19 $\pm$ 0.02 & 0.29 $\pm$ 0.04 & 0.19 $\pm$ 0.03 & 0.18 $\pm$ 0.03 \\
\cline{3-11}
7 & & \textbf{3 patch sizes} & 0.42 $\pm$ 0.32 & 0.90 $\pm$ 0.70 & 0.49 $\pm$ 0.20 & 0.23 $\pm$ 0.07 & 0.12 $\pm$ 0.01 & 0.29 $\pm$ 0.02 & 0.13 $\pm$ 0.02 & 0.10 $\pm$ 0.02\\\cline{3-11}
8 & & 3 patch sizes, slices & 0.33 $\pm$ 0.15 & 0.72 $\pm$ 0.37 & 0.55 $\pm$ 0.15 & 0.25 $\pm$ 0.08 & 0.12 $\pm$ 0.01 & 0.35 $\pm$ 0.04 & 0.11 $\pm$ 0.01 & 0.10 $\pm$ 0.02 \\
9 & & 3 patch sizes, ortho & 0.19 $\pm$ 0.05 & 0.65 $\pm$ 0.56 & 0.70 $\pm$ 0.38 & 0.21 $\pm$ 0.05 & 0.12 $\pm$ 0.01 & 0.25 $\pm$ 0.04 & 0.12 $\pm$ 0.02 & 0.10 $\pm$ 0.02\\\hline
10 & Young adults & \textbf{3 patch sizes} & 1.17 $\pm$ 0.80 & - & 0.75 $\pm$ 0.07 & 0.52 $\pm$ 0.12 & 0.21 $\pm$ 0.05 & 0.64 $\pm$ 0.28 & 0.28 $\pm$ 0.06 & - \\\cline{3-11}
11 & & 3 patch sizes, ortho & 0.65 $\pm$ 0.19 & - & 0.64 $\pm$ 0.20 & 0.77 $\pm$ 0.67 & 0.20 $\pm$ 0.05 & 0.52 $\pm$ 0.35 & 0.27 $\pm$ 0.06 & - \\\hline
\end{tabular}
\end{table*}

\section{Discussion}
\label{sec:discussion}
We have presented a method for the automatic segmentation of MR brain images, which has been evaluated on manually segmented preterm neonatal and adult images. Accurate segmentation results were obtained in images acquired at different ages (preterm neonatal vs. ageing adult) and with different acquisition protocols (coronal vs. axial, T\textsubscript{2}- vs. T\textsubscript{1}-weighted). The method uses CNNs to automatically extract the relevant information from the training data to learn convolution kernels, which allows adaptation to the problem at hand. Post-processing of the segmentation results, which is typically performed in brain image segmentation tasks, was here not necessary.

The method achieved accurate segmentations in terms of Dice coefficients for all tissue classes except mWM in the neonatal images. mWM consists, especially in the images acquired at 30 weeks PMA, of very few voxels, hence each mislabelled voxel strongly influences the Dice coefficient. Furthermore, this tissue is poorly visible in T\textsubscript{2}-weighted images and slightly better visible in T\textsubscript{1}-weighted images, which makes the manual annotation very difficult and prone to inter-observer variability \cite{Isgu15}. This consequently has influence on the performance of the neighbouring tissue classes, which can be seen from e.g. the performance on BS for the coronal images acquired at 40 weeks PMA (Table \ref{tab:diceresults}: Experiment 3). Even though the performance in terms of Dice coefficients is lowest for mWM, the location of the tissue class is quite well recognised by the method, even based on T\textsubscript{2}-weighted images only. This can be seen from the mean surface distances (Table \ref{tab:diceresults}) and from Figure \ref{fig:segmentations}: e.g. for the coronal images acquired at 40 weeks PMA, where mWM is recognised but does not follow the exact shape of the manual segmentation. 

The method uses patches of three different sizes. Because of the highly anisotropic voxel sizes in 4 of the 5 evaluated image sets, the use of orthogonal or 3D patches would result in anisotropic patches, non-square/cubic patches, or strongly resampled data. Therefore, only in-plane information was used, i.e. the volumetric images were segmented using information from the imaging plane only. This corresponds to the way a human observer performs annotation of these images: by selecting voxels that belong to each of the tissue types in a single image plane based on trained anatomical knowledge. This approach is further motivated by additional experiments using patches from consecutive slices and using (resampled) orthogonal patches (Table \ref{tab:diceseparatepatches}: Experiments 8, 9 and 11), which did not provide information that allowed more accurate classification. The performance in terms of average Dice coefficient over all tissue classes remained 0.87 for the anisotropic images acquired at 30 weeks PMA, and improved slightly, from 0.90 to 0.91, for the (nearly) isotropic images of young adults. The limited increase in performance for the images of young adults was possibly influenced by the manual segmentations, which have been performed in the coronal plane and are therefore less accurate in the other planes. Furthermore, experiments with interpolated 3D patches of single sizes suggested that 3D convolutions also do not necessarily result in increased performance (Table \ref{tab:diceseparatepatches}: Experiments 2 and 4), possibly influenced by the large increase in number of weights and biases in the optimisation. The performance of a multi-scale 3D CNN was not evaluated because training a network with three large 3D input patches would become computationally infeasible.

The results using each of the three patches separately show that each patch focusses on a different aspect of the segmentation problem. The smallest patch allows detailed analysis of local texture but misses spatial consistency. The largest patch results in a smooth segmentation but misses small details. This can especially be seen for the segmentation of cGM; the average Dice coefficients for each of the patch sizes separately is 0.78, while the combined patches result in an average Dice coefficient of 0.84 (Table \ref{tab:diceseparatepatches}: Experiments 1, 3, 5 and 7). This can also be observed in Figure \ref{fig:zoomedsegmentations}: the internal CSF in the lateral sulcus is recognised using only the smallest patch size, but not with the two larger sizes. On the other hand, spatially inconsistent results are obtained for the segmentation of the hippocampus using only the smallest patch size, whereas the largest patch size shows better consistency in that respect. In both cases, the result with the three patches combined shows the most accurate segmentation. The kernels and responses in the first layer (Figure \ref{fig:kernels}) show that different kernels enhance different parts of the image and the larger kernels operate on a larger scale. In addition, the proposed multi-scale approach is compared with the multi-scale approach of Sermanet and LeCun \cite{Serm11} and shows better performance for every tissue class.

The method is based on T\textsubscript{2}-weighted images for the neonatal images and on T\textsubscript{1}-weighted images in adults. Furthermore, the segmentation of both coronal and axial images was evaluated. This shows that the method is able to adapt to different acquisition protocols based on representative training data. Furthermore, being able to segment the images based on a single T\textsubscript{1}- or T\textsubscript{2}-weighted image, instead of relying on multiple acquisitions, allows omitting registration between images and thus precludes possible registration errors. 

The segmentation method was applied to MR images of the developing, the young adult, and the ageing brain. No images acquired between these three age ranges were evaluated, but the method can likely be straightforwardly applied to images acquired at other ages. The method might however be limited in the application to images with abnormalities that are not included in the training set.

Additional evaluation in the segmentation of 134 classes as provided in the MICCAI challenge on multi-atlas labelling demonstrated that the approach can straight-forwardly be applied to a different segmentation task, without any parameter tuning. This resulted in an overall average Dice coefficient of 0.7353 as compared with 0.725 reported by de Br\'{e}bisson et al. \cite{Breb15}.

A brain mask \cite{Smit02a} is used to limit the number of samples considered in the classification. Because the results of this brain mask were not always consistent for the images of ageing adults, a very conservative setting for the fractional intensity parameter was chosen for these images such that almost the whole head was included in the mask. It might be possible to completely omit the use of a brain mask, but this would also increase the computational load.

CNNs are known to need a large number of training samples. Because we applied CNNs in a voxel classification task, many training samples were available. They were, however, extracted from a limited number of training images. More training images could therefore increase the performance. In addition, including diverse training data from different data sets in a single training step could generalise the problem and would in this way not require retraining the network. This generalisability over a more diverse data set might, however, be achieved at the cost of a decreased performance on each of the data sets that are included.

Many adjustments to the architecture of the network are possible, including more or different patch and kernel sizes, a different number of convolution layers, or a different number of nodes or layers in the fully connected network. Several settings were evaluated in preliminary experiments, but future work could focus on additional optimisation in this respect. Note that the network architecture provides a search space for the method and all weights and biases are optimised by the system. Small variations in the architecture are therefore not likely to have a large effect on the performance.

\section{Conclusion}
\label{sec:conclusion}
The presented CNN method for the automatic segmentation of MR brain images shows accurate segmentation results in images acquired at different ages and with different acquisition protocols.

% or


%\appendices
%% you can choose not to have a title for an appendix
%% if you want by leaving the argument blank
%\section{}


\section*{Acknowledgment}
\label{sec:acknowledgement}
The authors gratefully acknowledge the support of NVIDIA Corporation with the donation of the Tesla K40 GPU used in this research. 

The authors thank Bennett Landman for providing the data of the MICCAI challenge on multi-atlas labelling.

%\ifCLASSOPTIONcaptionsoff
%  \newpage
%\fi



\bibliographystyle{ieeetr}
%\begin{thebibliography}{1}

\IEEEtriggeratref{34}
\bibliography{references}

%\end{thebibliography}


%\end{IEEEbiography}
%% insert where needed to balance the two columns on the last page with
%%\newpage


%\vfill


\end{document}